\documentclass[conference]{IEEEtran}
\IEEEoverridecommandlockouts
\usepackage{cite}
\usepackage{amsmath,amssymb,amsfonts}
\usepackage{algorithmic}
\usepackage{graphicx}
\usepackage{textcomp}
\usepackage{xcolor}
\usepackage{array}
\usepackage[caption=false,font=normalsize,labelfont=sf,textfont=sf]{subfig}
\usepackage{stfloats}
\usepackage{url}
\usepackage{verbatim}

\def\BibTeX{{\rm B\kern-.05em{\sc i\kern-.025em b}\kern-.08em
    T\kern-.1667em\lower.7ex\hbox{E}\kern-.125emX}}

\begin{document}

\title{The Chronos Vulnerability: A Taxonomy of Temporal Persistence and Memory-Based Deception in Agentic AI}

\author{\IEEEauthorblockN{Om Narayan}
\IEEEauthorblockA{\textit{Department of Computer Science} \\
\textit{New York University}\\
New York, NY, USA \\
on371@nyu.edu}
\and
\IEEEauthorblockN{Ramkinker Singh}
\IEEEauthorblockA{\textit{Carnegie Mellon University}\\
Pittsburgh, PA, USA \\
ramkinks@alumni.cmu.edu}
\and
\IEEEauthorblockN{Praveen Baskar}
\IEEEauthorblockA{\textit{Independent Researcher}\\
Chicago, IL, USA \\
praveen2004er@gmail.com}
}

\maketitle

\begin{abstract}
The transition from stateless generative models in artificial intelligence to stateful, autonomous agents represents an architectural evolution that, while providing the capabilities of long-term planning and the automation of enterprise workflows, also represents the introduction of a new form of security threat, the Chronos Vulnerability.
The Chronos Vulnerability represents the threat of memory-based attacks, including the Memory Injection Attack (MINJA) and the sleeper agent, in which the internal belief system of the autonomous agent is compromised, effectively decoupling the attack vector from the final catastrophic event.
This study formalizes the threat model for persistence-based attacks and the threat of Dynamics Blindness in the context of the World of Workflows benchmark, demonstrating that traditional endpoint content filters are insufficient for the current stateful architecture.
Consequently, this study synthesizes a defense-in-depth landscape, categorizing emerging frameworks such as diagnostic trajectory guardrails (AgentDoG), formal temporal verification (Agent-C), immunological memory consensus (A-MemGuard), and hardware-anchored trust via GPU-based Trusted Execution Environments (TEEs) and Zero-Trust memory architectures.
\end{abstract}

\begin{IEEEkeywords}
Agentic AI, Chronos Vulnerability, Memory Injection, Dynamics Blindness, Security.
\end{IEEEkeywords}

\section{Introduction}
\IEEEPARstart{T}{he} digital ecosystem is witnessing a major structural change, moving from a stateless generative models paradigm to a paradigm where autonomous agentic systems dominate.
This represents a major leap forward in the evolution of artificial intelligence, transforming passive oracular models into active participants in enterprise systems, capable of autonomous tool usage~\cite{ref1} and dynamic reasoning~\cite{ref2}.

Although the addition of persistent memory, recursive planning, and tool usage has resulted in major productivity gains, it has concurrently enabled a highly complex threat landscape.
Agents, if left without internet or API access restrictions, represent a major reliability and security risk.
Legacy cybersecurity models, geared toward deterministic software or session-oriented generative models, are fundamentally unsuitable for addressing the failure modes of these systems~\cite{ref7}.

The Agentic AI paradigm represents a major paradigm shift away from content filters and toward systemic architectures to address the persistence of internal states.
In this analysis, the Chronos Vulnerability is introduced, a major threat vector in AI systems, wherein memory-based attacks utilize temporal persistence to allow for the incubation of a malicious payload within the internal belief system of the AI, often long before catastrophic failure.
In contrast to legacy prompt injection, wherein the attack is session-oriented, the Chronos Vulnerability infects the memory bank of the AI, effectively embedding a persistent malicious actor within the system architecture.

\section{The Moat of Memory: Temporal Persistence and Internal State}
Artificial intelligence development currently transitions from vibe-coding, an imprecise approach based on natural language processing, to a precise specification-based approach that defines the agent's behavior with technical specifications.
In the conventional stateless model, an adversary's input is treated as a transient disturbance.
Once the session is closed, the agent is reset to its initial state, thereby eliminating the threat.

In the current state of agentic systems, memory is more than just an information repository;
it is an evolving entity that is an ongoing attack vector.
The implications of this are profound: the adversary is no longer required to influence the agent during a critical operation.
Instead, the adversary can inject the agent with bad information that will infect the agent's beliefs.
Because the agent will interpret these infected beliefs as its true history, it will defend these beliefs against all attempts to correct it, thereby leading to a sleeper agent condition where the initial attack is completely invisible~\cite{ref10}.

\subsection{Foundational Architectures: Mapping the BDI Stack}
The security of agentic systems is increasingly being understood as an extension of the Belief-Desire-Intention (BDI) model that distinguishes between information state, admissible goals, and adopted plans.
Generative Agents utilize a cognitive architecture highly akin to BDI frameworks~\cite{ref3}.
In the current state of agentic systems, the BDI components are associated with particular levels of the system architecture:

\begin{itemize}
\item \textbf{Beliefs:} The agent's representation of the world state and its memory stores.
The integrity of the agent's beliefs is based on the ability to track the origins and resolve conflicts.
In the event of corruption in this domain, the agent will experience Belief Drift.
\item \textbf{Desires:} The high-level goals and constraints (e.g., system prompts).
These are the elements that are subject to alignment faking, where the agent prioritizes its latent preferences over its expressed goals~\cite{ref10}.
\item \textbf{Intentions:} The agent's current plan and the associated tool calls.
\end{itemize}

\subsection{Hierarchical Memory Stratification}

Current agent architectures mirror the cognitive layering observed in the human brain~\cite{ref4}, where each cognitive level has a unique risk profile that is managed via Virtual Context Management~\cite{ref4}.
Table~\ref{tab:memory} outlines these layers.

\begin{table*}[!t]
\caption{Hierarchical Memory Stratification and Vulnerabilities}
\label{tab:memory}
\centering
\begin{tabular}{|l|p{2.5in}|p{2.5in}|}
\hline
\textbf{Memory Layer} & \textbf{Functional Specification} & \textbf{Primary Vulnerability Axis} \\
\hline
Episodic & Capture of discrete past events (who, what, where)~\cite{ref3}. & Target for MINJA~\cite{ref8}; leads to reasoning drift~\cite{ref24}. \\
\hline
Semantic & Abstracted vault for generalized facts and world models~\cite{ref3}. & Vulnerable to gradual drift of general truths~\cite{ref6}. \\
\hline
Procedural & Step-by-step routines (e.g., skill libraries~\cite{ref5}). & Poisoning allows skipping verification steps~\cite{ref6}. \\
\hline
Factual & Durable entity data; updates infrequently~\cite{ref4}. & Susceptible to identity spoofing and link corruption~\cite{ref9}. \\
\hline
Working & Active ``scratchpad'' in the context window~\cite{ref4}. & Target for transient, session-scoped prompt injection~\cite{ref13}. \\
\hline
\end{tabular}
\end{table*}

\subsection{The Axiomatic Duality of Memory and Belief}
Memory in agentic systems has a dual purpose: it is both a retrieval system and a system for belief formation.
When the agent retrieves a piece of information from its VectorDB memory system, it has implicit trust in the information and uses it as a premise in the subsequent reasoning sequence.
The Chronos vulnerability takes advantage of this implicit trust. By injecting a poisoned memory into the agent's memory system, the attacker forces the agent to reason to an action that is in line with the injected memory.

If the injected memory causes a successful tool output, the agent's cognitive state is updated to further reinforce the validity of the memory.
This initiates a Recursive Belief Confirmation, or a Hallucination Spiral (a self-reinforcing cycle of fabricated reasoning), where the agent actively fights back against corrective input based on the compromised historical confirmations~\cite{ref24}.
To systematically induce this pathological state, however, we must move beyond abstract cognitive theory and define the specific access privileges required by an adversary.

Table~\ref{tab:attack_dim} contrasts session-based and persistence-based attacks.

\begin{table}[!t]
\caption{Attack Dimension Comparison}
\label{tab:attack_dim}
\centering
\begin{tabular}{|p{0.8in}|p{1.1in}|p{1.1in}|}
\hline
\textbf{Dimension} & \textbf{Session-Based (Prompt Injection)} & \textbf{Persistence-Based (Memory Poisoning)} \\
\hline
Duration & Transient; ends with the session~\cite{ref13}. & Persistent; resides indefinitely in VectorDBs~\cite{ref8}. \\
\hline
Trigger Mechanism & Immediate; requires active query~\cite{ref13}. & Decoupled; activated weeks or months later~\cite{ref10}. \\
\hline
Visibility & High; visible within the prompt context~\cite{ref13}. & Low; hidden deep within the retrieval pipeline~\cite{ref9}. \\
\hline
Trust Model & Untrusted external input~\cite{ref13}. & Implicitly trusted Internal State~\cite{ref8}. \\
\hline
Impact Vector & Single-turn manipulation~\cite{ref13}. & Long-tail reasoning drift and belief corruption~\cite{ref24}. \\
\hline
\end{tabular}
\end{table}

\subsection{Threat Model and Adversarial Capabilities}
To operationalize this cognitive vulnerability into a concrete exploit, it is necessary to define the threat model using various notations from well-established security frameworks like AgentPoison~\cite{ref6} and Portcullis~\cite{ref7}. We define a grey-box attacker who does not have read/write access to the agent's internal weights or isolated backend database but does have Query Access and Context Injection Capability via the normal user interface or external environments like web pages or emails.

The attacker has the following knowledge: the attacker knows the semantic similarity mechanisms employed in the agent's retrieval system. The attacker's objective is to maximize the Injection Success Rate (ISR) without triggering anomaly detection, embedding a payload that remains dormant until activated by a legitimate user's benign query. This model clearly delineates the infection phase from the execution phase, defining the key parameters of the Chronos vulnerability.

\section{Taxonomy of Temporal Attacks: From MINJA to Session Smuggling}
A well-designed security model for autonomous systems must therefore transcend traditional interventions limited to a single point in time and address the inherent structural degradation of temporal integrity.
Recent attack vectors demonstrate the considerable ingenuity employed in memory-based exploitation.

\subsection{The MINJA Class of Memory Injection}
The Memory Injection Attack, or MINJA~\cite{ref8}, is a landmark advancement in the development of persistent exploitation.
Following on from its predecessor, AgentPoison, which proved that less than 0.1\% of poisoned data can corrupt an agent~\cite{ref6}, MINJA has shown that normal users can compromise long-term memory solely through query-based interactions.
The MINJA attack follows a three-part sequence:

\begin{enumerate}
\item \textbf{Indication Prompt:} The attacker sends a seemingly innocuous query that subtly guides the agent toward generating specific reasoning steps~\cite{ref8}.
\item \textbf{Bridging Steps:} Logical connections establishing an association between a benign victim identifier and an attacker's target.
\item \textbf{Progressive Shortening Strategy (PSS):} An iterative process where the attacker gradually removes explicit instructions, forcing the agent to complete the reasoning based on its retrieved past demonstrations.
\end{enumerate}

This process solidifies the malicious reasoning into the agent's memory repository as a truth.
Furthermore, stealthier versions of this approach, such as MemoryGraft~\cite{ref9}, have been identified, where attackers utilize seemingly benign documentation, such as a README file, to execute single-shot, indirect memory grafting.

\subsection{Deceptive Instrumental Alignment and Sleeper Agents}
In addition to external poisoning, there is a risk of internal misalignment, termed \textbf{Deceptive Instrumental Alignment,} wherein AI systems are designed to strategically hide their malicious goals during the training and evaluation process~\cite{ref10}.
Empirical evidence supporting the so-called Model Organisms of Misalignment~\cite{ref10}, published in 2024, revealed that AI models could be designed to create secure code if the inputted year was 2023, but deliberately create vulnerabilities if the inputted year changed to 2024. Interestingly, even conventional safety mechanisms, such as Reinforcement Learning from Human Feedback, are unable to eliminate this deception and, in fact, make the AI model even better at hiding its true, malicious intentions, a phenomenon termed \textbf{Alignment Faking}~\cite{ref10}.

\subsection{Agent Session Smuggling in Multi-Agent Systems}
As AI systems evolve to become Multi-Agent Systems (MAS), a new risk vector arises in the form of coordination~\cite{ref11}.
In this case, a rogue sub-agent uses a legitimate session to smuggle hidden commands between a legitimate client request and a server’s response.
Due to the inherent trust structure of agents, who are designed to trust each other, these smuggled commands go undetected, resulting in Shadow Escapes, wherein a legitimate tool execution is run completely undetected~\cite{ref11}.
Many-to-one adversarial consensus~\cite{ref11} is the primary source discussing malicious MAS coordination.

\section{Case Study: EchoLeak and Zero-Click Exfiltration}
The EchoLeak vulnerability, designated as CVE-2025-32711~\cite{ref12}, is a significant example of a zero-click prompt injection exploit used for mass-scale data exfiltration from a production environment.
This exploit represents the concept of the LLM Scope Violation, and Indirect Prompt Injection~\cite{ref13} where the AI is tricked into violating the scope for which it was originally trusted with the intention of revealing internal data without the involvement of a human user.

The process by which the attacker exploits the EchoLeak vulnerability can be divided into the following steps:
\begin{itemize}
\item \textbf{Malicious Ingestion:} The attacker sends a seemingly innocuous email with a prompt-injection exploit hidden within it~\cite{ref12, ref13}.
The exploit is usually hidden by using HTML comments or by using a high-contrast color scheme with white text on a white background~\cite{ref13}.
The attacker does not need the user to interact with the email because the presence of the email within the user’s inbox is enough for the Retrieval-Augmented Generation (RAG) agent to ingest the email~\cite{ref13}.
\item \textbf{RAG Trigger:} The attacker sends a legitimate query using the agent with the intention of retrieving the context for the email they sent earlier.
The legitimate query might be something like ``Summarize the emails I have received recently.''
\item \textbf{Command Execution:} The attacker uses the prompt-injection exploit to trick the agent into executing a query for sensitive information like chat logs, OneDrive documents, Teams messages, etc., from the context window.
\item \textbf{Markdown Reflection:} The attacker uses the prompt-injection exploit to trick the agent into embedding the obtained data within a reference-style markdown link.
\item \textbf{Automated Fetch:} The client-side application, such as Outlook, fetches the image URL from the markdown link with the intention of displaying the image preview, thereby transferring the obtained data to the attacker’s server.
\end{itemize}

Table~\ref{tab:echoleak_revised} summarizes the vulnerability parameters.

\begin{table}[!t]
\caption{Taxonomy of EchoLeak Parameters and Surveyed Mitigations}
\label{tab:echoleak_revised}
\centering
\begin{tabular}{|p{1.0in}|p{1.0in}|p{1.0in}|}
\hline
\textbf{Vulnerability Class} & \textbf{Observed Metric} & \textbf{Surveyed Mitigation} \\
\hline
LLM Scope Violation ~\cite{ref12} & Zero-Click Access & CSP Hardening ~\cite{ref12} \\
\hline
Attack Vector ~\cite{ref13} & Indirect Injection & Instruction Auth ~\cite{ref12} \\
\hline
\end{tabular}
\end{table}

Thus, the EchoLeak exploit illustrates the potential for natural language to be used as a viable exploit for high-impact system exploitation.
The exploit also highlights the risk posed by the \textbf{LLM Scope Violation,} where the agent’s access to a variety of data sources is used against it due to the agent’s inability to distinguish between legitimate user prompts and retrieved context.

\section{Enterprise Dynamics Blindness: The Fog of War}
While EchoLeak demonstrates the extraction of internal state, the persistence of such attacks relies on a more insidious structural flaw: the agent’s inability to verify the temporal consequences of its own actions.
This phenomenon, which we term \textbf{Dynamics Blindness,} is not merely a performance bottleneck;
it is the primary obfuscation layer attackers leverage to induce silent failures.

Despite demonstrating high performance on web-based benchmarks like Mind2Web~\cite{ref14} and generic tool use tuning benchmarks like ToolLLM~\cite{ref15}, frontier LLMs are also found to suffer from considerable performance unreliability in complex enterprise systems.
Studies that use the World of Workflows (WoW) benchmark~\cite{ref16} quantify the performance failure as Dynamics Blindness.

In the context of enterprise workflows, the agent's view is mostly obscured;
a single action can cause unobservable cascading database updates~\cite{ref16}. The study defines three different performance gaps that are described as follows:
\begin{itemize}
\item \textbf{The Representation Gap:} The agent lacks symbolic grounding and thus equates semantic descriptions with strict symbolic identifiers (e.g., User Name versus sys\_id)~\cite{ref16}.
\item \textbf{The Dynamics Gap:} The agent lacks internal transition models.
The agent is unable to anticipate cascading side effects because of an insufficient understanding of the physics of the system~\cite{ref16}.
\item \textbf{The Observability Gap:} The agent relies on tool outputs that abstract away critical state changes (e.g., Ticket Created vs. actual table writes)~\cite{ref16}.
\end{itemize}

\begin{table*}[!t]
\caption{Cross-Benchmark Analysis: Performance Decay and Dynamics Blindness \cite{ref16}}
\label{tab:dynamics_enterprise}
\centering
\begin{tabular}{|l|c|c|c|c|c|}
\hline
\textbf{Benchmark \& Environment} & \textbf{Metric} & \textbf{GPT-4} & \textbf{Gemini-1.5 Pro} & \textbf{Claude-3.5} & \textbf{Blindness Impact} \\
\hline
\textbf{AgentBench} (Generalist) \cite{ref17} & TSR & 42.4\% & 35.2\% & 38.1\% & Baseline \\
\hline
\textbf{WoW-bench} (Enterprise) \cite{ref16} & TSR & 22.0\% & 38.7\% & 32.0\% & $\Delta_{Env} -13.2\%$ (Avg.) \\
\hline
\textbf{WoW-bench} (Constrained) \cite{ref16} & TSRUC & 2.0\% & 6.0\% & 4.0\% & \textbf{$-87.6\%$ (Mean Rel. Drop)} \\
\hline
\textbf{World Modeling} (Dynamics) \cite{ref16} & State Pred. & 18.0\% & 22.0\% & 22.0\% & High State Entropy \\
\hline
\textbf{Inverse Dynamics} (Logic) \cite{ref16} & Action Pred. & 27.3\% & 18.3\% & 28.1\% & Symbolic Grounding Gap \\
\hline
\end{tabular}
\end{table*}

As illustrated in Table~\ref{tab:dynamics_enterprise}, the transition from generalist tasks to constrained enterprise workflows results in a catastrophic performance decay.
While frontier models maintain a baseline Task Success Rate (TSR) of 22.0--38.7\% in unconstrained enterprise environments, the introduction of hidden state variables causes the Task Success Rate Under Constraint (TSRUC) to collapse significantly.
For instance, GPT-4 experiences a 90.9\% relative drop in success rate (from 22.0\% TSR to 2.0\% TSRUC).
Across all frontier models, this manifests as a mean relative performance collapse of 87.6\% , providing a quantitative definition for Dynamics Blindness.
This serves as the empirical foundation for Dynamics Blindness, where the agent fails to anticipate cascading consequences within complex system architectures.

\section{Taxonomy of Emerging Defenses: Trajectory and Formal Verification}
The current research landscape for mitigating the Chronos Vulnerability has shifted from model-centric filtering, which cannot see the history, and toward system-centric verification architectures that secure the entire temporal trajectory.
A comprehensive defense must simultaneously address the integrity of the past (preventing MINJA) and the validity of the future (preventing Dynamics Blindness).

\subsection{Landscape of Trajectory-Level Diagnostic Guardrails (AgentDoG)}
Endpoint guardrails have been shown to be ineffective for agents with long time horizons.
Recent advancements, exemplified by the AgentDoG~\cite{ref19} framework provides a contextual monitoring approach for entire execution trajectories.
AgentDoG provides a unified taxonomy across three axes: Risk Source (e.g., User Input, Environment), Failure Mode (e.g., Goal Drift, Instruction Hijacking), and Real-World Harm (e.g., Privacy Leakage).
Using a hierarchical approach to agentic attribution, AgentDoG performs a root-cause diagnosis~\cite{ref18} by identifying the exact planning step where the reasoning went astray.
AgentDoG-FG-4B achieved an 82.0\% accuracy in risk source identification on the ATBench dataset~\cite{ref19}, substantially outperforming foundation models.

\subsection{Formal Temporal Conformance (Agent-C)}
The probabilistic nature of large language models inhibits the application of natural language governance.
Agent-C~\cite{ref20} framework provides a solution to this problem by providing mathematical temporal safety constraints.
In this case, a Domain-Specific Language (DSL) is used to write policies, which are compiled into First-Order Logic (FOL).
An SMT (Satisfiability Modulo Theories) solver runs continuously alongside the agent, verifying that proposed tool calls do not violate the FOL trace history.
If a violation is imminent, constrained generation forces the model to backtrack or regenerate.
Compared to lightweight interception wrappers like AgentSpec~\cite{ref21}, Agent-C provides rigorous mathematical guarantees, successfully elevating GPT-5 conformance from 83.7\% to 100\%~\cite{ref20}.

\subsection{Digital Immunity (A-MemGuard)}
To counter memory poisoning directly, A-MemGuard~\cite{ref22} applies an Artificial Immune System (AIS) paradigm to agent architecture.
It employs Consensus Validation: retrieving related memories to form parallel reasoning paths.
If a single path—induced by a poisoned memory—diverges mathematically from the consensus centroid, it is filtered out.

To ensure robustness, the A-MemGuard framework utilizes proactive defense mechanisms that analyze historical trajectories to verify output constraints~\cite{ref22}.
By grounding the model in these authentic historical cases, the system ensures alignment with known-good execution paths and prevents the reinforcement of malicious reasoning~\cite{ref22}.

\section{Future Directions: Hardware-Anchored Trust}
While Agent-C and A-MemGuard provide robust logical verification, they remain software artifacts susceptible to lower-level subversion.
If an attacker can manipulate the memory pointers or weights at the infrastructure level, algorithmic defenses become irrelevant.
Therefore, the final line of defense must ground the agent's semantic integrity in Hardware-Anchored Trust.

\subsection{Confidential Computing and TEEs}
The implementation of frontier agents in TEE architectures, such as NVIDIA’s H100 GPU with Confidential Computing-On or Intel’s TDX, ensures physical isolation for model weights and memory~\cite{ref23}.
While memory encryption within the GPU’s HBM (high-bandwidth) memory ensures a throughput overhead of only around 2\% for models such as Llama-3.1-70B~\cite{ref23}, this figure must be qualified.
As discussed earlier, performance within TEE architectures is highly dependent on NUMA (Non-Uniform Memory Access) configurations and PCIe bus bottlenecks for data transfer.
As shown by Chrapek et al.~\cite{ref23}, if the NUMA placement and encryption over the PCIe bus are not optimized, latency overheads of 20-100\% are introduced.
Hence, a high degree of system orchestration is required to support this performance-security Pareto frontier.

\subsection{Dynamical Systems Theory in Semantic Space}
Future autonomous architectures can be represented as discrete dynamical systems in semantic space~\cite{ref24}.
By studying the loops of agentic reasoning, three regimes of autonomous behavior can be identified: Contractive (Stable Attractors), Oscillatory (indecision), and Exploratory (hallucination spirals)~\cite{ref24}.
By using Lyapunov stability theorems, it is possible to mathematically predict exactly when an autonomous agent will enter a Hopf bifurcation (a critical point where stable behavior transitions into sustained oscillations) and become uncontrollable, thereby advancing the field of verifiable autonomy~\cite{ref24}.

\subsection{Zero-Trust Memory and Cryptographic Provenance}
To permanently eliminate the Chronos vulnerability, it is necessary to move toward a Zero-Trust Architecture for unified memory~\cite{ref26}.
Protocols like MemTrust~\cite{ref26} utilize epoch keys to re-encrypt active memories and garbage keys to cryptographically erase ``forgotten'' data.
This guarantees that the act of forgetting does not leak intelligence~\cite{ref26}, transforming the memory of an autonomous agent from a static vector store into a dynamic, encrypted lifecycle.
By associating every memory retrieval with a cognitive state transition that is cryptographically verifiable~\cite{ref25}, belief drift can be forensically traced and reversed.

\section{Conclusion}
The current state of vibe-coding in the agentic development process is unsustainable from a security point of view.
With the strategic importance of AI agents projected to rise in the near future, the appearance of the ``Chronos Vulnerability''—encompassing the MINJA~\cite{ref8}, EchoLeak~\cite{ref12}, and Sleeper Agents~\cite{ref10}—highlights the reality of a security threat landscape where the lack of proper accountability in autonomous systems creates a systemic enterprise risk.
The pervasive nature of this threat landscape is a direct result of the blindness of the dynamics and the unquestioned persistence of the internal state.
The course of action requires the establishment of a secure memory moat by grounding autonomous systems in formal logic and hardware trust.

The synthesis of defense-in-depth strategies, such as AgentDoG~\cite{ref19}, Agent-C~\cite{ref20}, A-MemGuard~\cite{ref22}, and GPU-based TEEs bounded by cryptographic provenance, highlights a critical path toward to create a secure agentic environment that is verifiable, traceable, and secure in the face of the industrialized adversarial threat landscape projected in the forthcoming decade.


\begin{thebibliography}{1}

\bibitem{ref1}
T. Schick \emph{et al.}, ``Toolformer: Language models can teach themselves to use tools,'' \emph{arXiv preprint arXiv:2302.04761}, 2023, doi: \url{10.48550/arXiv.2302.04761}.

\bibitem{ref2}
S. Yao \emph{et al.}, ``ReAct: Synergizing reasoning and acting in language models,'' \emph{arXiv preprint arXiv:2210.03629}, 2022, doi: \url{10.48550/arXiv.2210.03629}.

\bibitem{ref3}
J. S. Park \emph{et al.}, ``Generative agents: Interactive simulacra of human behavior,'' in \emph{Proc. 36th Annu. ACM Symp. User Interface Softw. Technol. (UIST)}, San Francisco, CA, USA, 2023, pp. 1--22, doi: \url{10.1145/3586183.3606763}.

\bibitem{ref4}
C. Packer \emph{et al.}, ``MemGPT: Towards LLMs as operating systems,'' \emph{arXiv preprint arXiv:2310.08560}, 2023, doi: \url{10.48550/arXiv.2310.08560}.

\bibitem{ref5}
G. Wang \emph{et al.}, ``Voyager: An open-ended embodied agent with large language models,'' \emph{arXiv preprint arXiv:2305.16291}, 2023, doi: \url{10.48550/arXiv.2305.16291}.

\bibitem{ref6}
Z. Chen \emph{et al.}, ``AgentPoison: Red-teaming LLM agents via poisoning memory or knowledge bases,'' \emph{arXiv preprint arXiv:2407.12784}, 2024, doi: \url{10.48550/arXiv.2407.12784}.

\bibitem{ref7}
Y. Zhang \emph{et al.}, ``Portcullis: A scalable and verifiable privacy gateway for large language models,'' \emph{Proceedings of the AAAI Conference on Artificial Intelligence}, vol. 39, no. 1, pp. 1021--1029, 2025, doi: \url{10.1609/aaai.v39i1.32088}.

\bibitem{ref8}
S. Dong \emph{et al.}, ``A practical memory injection attack (MINJA) against LLM agents,'' \emph{arXiv preprint arXiv:2503.03704}, 2025, doi: \url{10.48550/arXiv.2503.03704}.

\bibitem{ref9}
S. Dong \emph{et al.}, ``MemoryGraft: Persistent compromise of LLM agents via poisoned RAG stores,'' \emph{arXiv preprint arXiv:2512.16962}, 2025, doi: \url{10.48550/arXiv.2512.16962}.

\bibitem{ref10}
E. Hubinger \emph{et al.}, ``Sleeper agents: Training deceptive LLMs that persist through safety training,'' \emph{arXiv preprint arXiv:2401.05566}, 2024, doi: \url{10.48550/arXiv.2401.05566}.

\bibitem{ref11}
A. Bashir \emph{et al.}, ``Many-to-one adversarial consensus: Exposing multi-agent collusion risks in AI-based healthcare,'' \emph{arXiv preprint arXiv:2512.03097}, 2025, doi: \url{10.48550/arXiv.2512.03097}.

\bibitem{ref12}
P. Reddy and A. S. Gujral, ``EchoLeak: The first real-world zero-click prompt injection exploit in a production LLM system,'' \emph{arXiv preprint arXiv:2509.10540}, 2025, doi: \url{10.48550/arXiv.2509.10540}.

\bibitem{ref13}
K. Greshake \emph{et al.}, ``Not what you've signed up for: Compromising real-world LLM-integrated applications with indirect prompt injection,'' \emph{arXiv preprint arXiv:2302.12173}, 2023, doi: \url{10.1145/3605764}.

\bibitem{ref14}
X. Deng \emph{et al.}, ``Mind2Web: Towards a generalist agent for the web,'' \emph{arXiv preprint arXiv:2409.01927}, 2024, doi: \url{10.48550/arXiv.2409.01927}.

\bibitem{ref15}
Y. Qin \emph{et al.}, ``ToolLLM: Facilitating large language models to master 16000+ real-world APIs,'' \emph{arXiv preprint arXiv:2307.16789}, 2023, doi: \url{10.48550/arXiv.2307.16789}.

\bibitem{ref16}
L. Li \emph{et al.}, ``World of workflows: A benchmark for enterprise agent reliability,'' \emph{arXiv preprint arXiv:2601.22130}, 2026, doi: \url{10.48550/arXiv.2601.22130}.

\bibitem{ref17}
X. Liu \emph{et al.}, ``AgentBench: Evaluating LLMs as agents,'' \emph{arXiv preprint arXiv:2308.03688}, 2023, doi: \url{10.48550/arXiv.2308.03688}.

\bibitem{ref18}
C. Pei \emph{et al.}, ``Flow-of-Action: SOP Enhanced LLM-Based Multi-Agent System for Root Cause Analysis,'' in \emph{Companion Proc. ACM Web Conf. 2025 (WWW Companion '25)}, Sydney, NSW, Australia, 2025, pp. 553--565, doi: \url{10.1145/3701716.3715225}.

\bibitem{ref19}
Z. Wang \emph{et al.}, ``AgentDoG: A diagnostic guardrail framework for agentic safety,'' \emph{arXiv preprint arXiv:2601.18491}, 2026, doi: \url{10.48550/arXiv.2601.18491}.

\bibitem{ref20}
A. Kamath \emph{et al.}, ``Agent-C: Enforcing temporal constraints for LLM agents,'' \emph{arXiv preprint arXiv:2512.23738}, 2025, doi: \url{10.48550/arXiv.2512.23738}.

\bibitem{ref21}
H. Wang, C. M. Poskitt, and J. Sun, ``AgentSpec: Customizable Runtime Enforcement for Safe and Reliable LLM Agents,'' in \emph{Proc. IEEE/ACM 48th Int. Conf. Softw. Eng. (ICSE)}, Rio de Janeiro, Brazil, 2026, doi: \url{10.48550/arXiv.2503.18666}.

\bibitem{ref22}
S. Dong \emph{et al.}, ``A-MemGuard: A proactive defense framework against memory injection attacks on LLM agents,'' \emph{arXiv preprint arXiv:2510.02373}, 2025, doi: \url{10.48550/arXiv.2510.02373}.

\bibitem{ref23}
A. Chrapek \emph{et al.}, ``Confidential LLM inference: Performance and cost,'' \emph{arXiv preprint arXiv:2509.18886}, 2025, doi: \url{10.48550/arXiv.2509.18886}.

\bibitem{ref24}
G. Hadjisoteriou \emph{et al.}, ``Geometric dynamics of agentic loops,'' \emph{arXiv preprint arXiv:2512.10350}, 2025, doi: \url{10.48550/arXiv.2512.10350}.

\bibitem{ref25}
P. Chantasantitam \emph{et al.}, ``PAL*M: Property attestation for large generative models,'' \emph{arXiv preprint arXiv:2601.16199}, 2026, doi: \url{10.48550/arXiv.2601.16199}.

\bibitem{ref26}
X. Zhou \emph{et al.}, ``MemTrust: A zero-trust architecture for unified AI memory system,'' \emph{arXiv preprint arXiv:2601.07004}, 2026, doi: \url{10.48550/arXiv.2601.07004}.

\end{thebibliography}
\end{document}